\documentclass[11pt]{article}
\usepackage{ifthen}
\usepackage{mdwlist}
\usepackage{amsmath,amssymb,amsfonts,amsthm}
\usepackage{bm}
\usepackage{bbm}

\usepackage[colorlinks=true,linkcolor=blue,citecolor=blue,urlcolor=blue]{hyperref}
\usepackage{enumitem}
\usepackage{graphicx}
\usepackage{xspace}
\usepackage{verbatim}
\usepackage{algorithm}
\usepackage{algpseudocode}
\usepackage[margin=1in]{geometry}
\usepackage{color}
\usepackage{thm-restate}
\usepackage{latexsym}
\usepackage{epsfig}
\usepackage[colorinlistoftodos,textsize=scriptsize]{todonotes}
\def\withcolors{0}
\def\withnotes{0}
\def\arxivver{1}

\renewcommand{\epsilon}{\ve}
\def\ve{\varepsilon}

\newcommand{\E}{\mbox{\bf E}}

\newcommand{\pr}[2][]{\mathrm{Pr}\ifthenelse{\not\equal{}{#1}}{_{#1}}{}\!\left[#2\right]}
\newcommand{\R}{\mathbb{R}}

\newcommand{\dtv}{d_{\mathrm {TV}}}

\newcommand{\normal}{\mathcal{N}}

\newcommand{\cD}{\mathcal{D}}

\newtheorem{theorem}{Theorem}

\newtheorem{proposition}[theorem]{Proposition}

\newtheorem{lemma}[theorem]{Lemma}

\newtheorem{definition}[theorem]{Definition}

\numberwithin{theorem}{section} 
\numberwithin{nontheorem}{section} 
\numberwithin{proposition}{section} 
\numberwithin{observation}{section} 
\numberwithin{remark}{section} 
\numberwithin{fact}{section} 
\numberwithin{lemma}{section} 
\numberwithin{claim}{section} 
\numberwithin{corollary}{section} 
\numberwithin{case}{section} 
\numberwithin{dfn}{section} 
\numberwithin{definition}{section} 
\numberwithin{question}{section} 
\numberwithin{openquestion}{section} 
\numberwithin{res}{section}

\ifnum\withcolors=1
  \newcommand{\new}[1]{{\color{red} {#1}}} % new
  \newcommand{\gcolor}[1]{{\color{red}#1}} 
\else
  \newcommand{\new}[1]{{{#1}}}

  \newcommand{\gcolor}[1]{{#1}}
\fi

\ifnum\withnotes=1
    
  \newcommand{\gnote}[1]{\par\gcolor{\textbf{G: }\sf #1}} % Gautam
  \newcommand{\gfootnote}[1]{\footnote{{\bf \gcolor{Gautam}}: {#1}}}

\else
  \newcommand{\gnote}[1]{}
  \newcommand{\gfootnote}[1]{}

\fi
\newcommand{\ignore}[1]{\leavevmode\unskip} % eat unnecessary spaces before

\title{The Broader Landscape of Robustness in Algorithmic Statistics}

\author {
Gautam Kamath\thanks{Cheriton School of Computer Science, University of Waterloo and Vector Institute. {\tt g@csail.mit.edu}.}}

\begin{document}
\maketitle

\begin{abstract}
The last decade has seen a number of advances in computationally efficient algorithms for statistical methods subject to robustness constraints. 
An estimator may be robust in a number of different ways: to contamination of the dataset, to heavy-tailed data, or in the sense that it preserves privacy of the dataset.
We survey recent results in these areas with a focus on the problem of mean estimation, drawing technical and conceptual connections between the various forms of robustness, showing that the same underlying algorithmic ideas lead to computationally efficient estimators in all these settings.
\end{abstract}

\section{Introduction}
\label{sec:intro}

Mean estimation is one of the most fundamental statistical tasks: given samples from a probability distribution, output an estimate of that distribution's mean. 
It is the prototypical question in statistical inference, and an important primitive that underlies a variety of more complex procedures (e.g., gradient descent, linear regression, etc.).
In other words, understanding mean estimation is a prerequisite for understanding essentially any other inference task.
As we will see, even in this basic setting, introducing new constraints or desiderata significantly affects the algorithmic ideas needed to solve the problem, particularly with a focus on computational efficiency.

More precisely, mean estimation refers to the following statistical problem. Given a dataset $X_1, \dots, X_n$ sampled i.i.d.\ from a distribution $\cD$ supported on $\R^d$, output an estimate $\hat \mu$ of its mean $\mu \triangleq \E_{X \sim \cD}[X]$.
The quality of the estimate is usually measured in terms of the $\ell_2$-distance between the true mean $\mu$ and the estimate $\hat \mu$.
We will focus on the following two cases for the distribution $\cD$:
\begin{enumerate}
    \item When it is a Gaussian distribution with identity covariance, i.e., $\cD = \normal(\mu, I)$; and 
    \item When it has bounded covariance, i.e., $\Sigma(\cD) \preceq I$.
\end{enumerate}
In either of these two cases, a textbook calculation demonstrates the efficacy of the empirical mean $\frac1n \sum_{i=1}^n X_i$. Specifically, we have that, with probability $\geq 95\%$,
\[\left\|\frac1n \sum_{i=1}^n X_i - \mu \right\|_2 \leq O\left(\sqrt{\frac{d}{n}}\right).\footnote{\new{For background on high-dimensional probability and statistics, consult the textbooks by Vershynin~\cite{Vershynin18} and Wainwright~\cite{Wainwright19}.}}\]
Furthermore, standard minimax lower bounds establish that no estimator can achieve a rate better than $\Omega\left(\sqrt{\frac{d}{n}}\right)$, demonstrating that this result is optimal up to constant factors.
To summarize, under fairly mild assumptions (i.e., only a bound on the variance of the distribution $\cD$), the most basic statistic possible (the empirical mean) achieves the optimal error rate for mean estimation.

However, the picture changes dramatically when we want our estimator to satisfy some flavor of \emph{robustness}.
Robustness can mean many different things.
We may want our estimator to be robust to \emph{contamination} of the distribution $\cD$.
We may desire that our estimator is robust in the case when the distribution $\cD$ has \emph{heavy tails}.
And we may require the estimator to preserve \emph{privacy} of the given dataset, particularly when the data represents sensitive information pertaining to individuals -- this too can be viewed as a form of robustness, as the estimator is not allowed to depend too much on individual data points.
As we will see, under any of these forms of robustness, the empirical mean is no longer the ideal solution, and the problem becomes substantially more complex.

\begin{table*}
\centering
\new{
\caption{Summary of different types of robustness considered in this article.}
\resizebox{\textwidth}{!}{\begin{tabular}{|c | c | c | c|} 
 \hline
 Type of Robustness & Input data  & Estimator's Goal & Section\\ 
 \hline
 Contamination & Stochastic, a fraction is modified arbitrarily & Bounded error, depending on corruption rate & Section~\ref{sec:robust} \\
 \hline
 Heavy-Tailed Data & Stochastic, from a distribution with heavy tails & Mild dependence of error on (inverse) failure probability & Section~\ref{sec:subgaussian} \\
 \hline
 Privacy & Arbitrary, a single point is modified arbitrarily & Approximately preserve estimator's output distribution & Section~\ref{sec:privacy} \\
 \hline
\end{tabular}}
}
\end{table*}

Over the last decade, there have been significant algorithmic advances on statistical estimation in these settings. 
While previous estimators ran into computational barriers, making them intractable for settings of even moderate dimensionality, these new results have produced the first computationally efficient algorithms for robust estimation.
We will survey these results with a focus on mean estimation.
Perhaps surprisingly, we will see that many of the same technical ideas and solution concepts underpin all of these (seemingly different) types of robustness, hinting at a broader theory for multivariate algorithmic statistics.
\if\arxivver0
\begin{figure}[h]
\centering
\includegraphics[width=7cm]{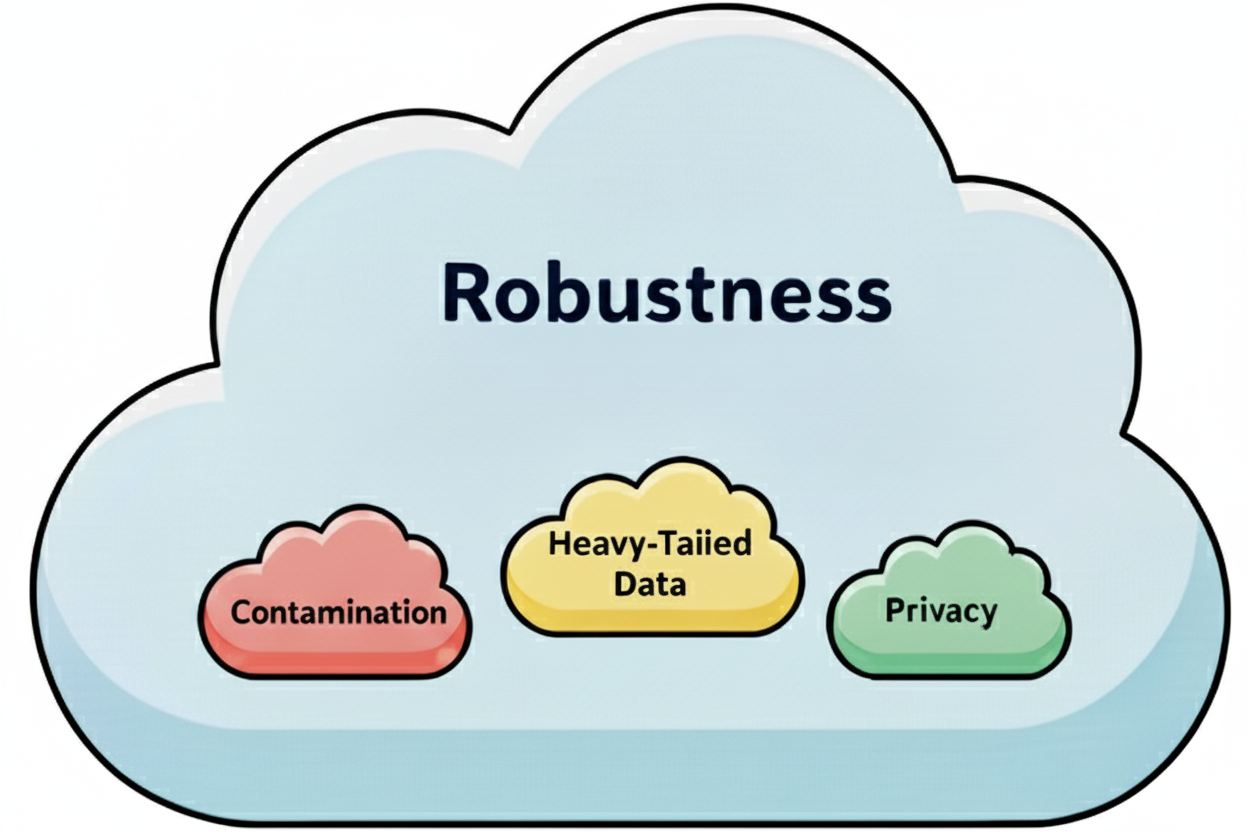}
\caption{The broader landscape of robustness contains considerations that include contamination, heavy-tailed data, and data privacy.}
\end{figure}
\fi
\section{Contamination}
\label{sec:robust}

We will first explore the most common notion of robustness for estimation tasks, robustness to \emph{contamination}.\footnote{In the literature, this setting is most commonly referred to as \emph{robust estimation}. However, to avoid confusion with the other forms of robustness, we will instead use the term \emph{contamination-robust estimation}.}
Our discussion so far relies heavily upon the assumption that samples are drawn i.i.d.\ from a well-behaved distribution $\cD$.
However, there are many reasons why this may not be the case for real-world data, and we can view our data as \emph{contaminated}.
By this, we mean that a fraction of our data comes from some other source, potentially adversarial in nature.
%For example, such a setting can capture the common case of model misspecification: while it is common to assume data is drawn from a Gaussian, this is a modeling assumption to \emph{approximate} the true data distribution -- nature rarely samples \emph{exactly} from Gaussians.
Reasons for this contamination can range from innocuous (e.g., errors by the data curator, model misspecification) to malicious (e.g., data poisoning attacks~\cite{SteinhardtKL17,DiakonikolasKKLSS19,CinaGDVZMOBPR23}).
There are a variety of ways this type of contamination could cause the data to diverge from our assumptions, and we would like our algorithms to be robust to such deviations.
%\subsection{The Contamination Model}

A classic and well-studied way to capture this style of robustness is Huber's contamination model~\cite{Huber64} from Statistics.
In this model, $(1-\eta)n$ samples are drawn i.i.d.\ from a distribution $\cD$, and $\eta n$ samples are drawn i.i.d.\ from some other (adversarially chosen) distribution.
The resulting dataset is thus drawn from a mixture of the original distribution and another distribution with $\eta$-mixing weight, and can be viewed as simulating an adversary who can add a small fraction of data to the dataset.

We will focus on an even stronger adversary, known as the \emph{$\eta$-strong contamination model}.
In this model, $n$ samples are first drawn i.i.d.\ from a distribution $\cD$. An adversary is allowed to inspect these samples, and replace any $\eta n$ of them with arbitrary other datapoints.\footnote{A common question is whether the algorithm has knowledge of the corruption fraction $\eta$. \new{In most cases,} this turns out to not matter much~\cite{JainOR22} \new{(though there are exceptions, e.g., for identifiability reasons when both $\eta$ and the covariance matrix are unknown, see Exercise 1.7 (b) of~\cite{DiakonikolasK22})}. For simplicity, we will assume $\eta$ is known.} 
The resulting (modified) dataset is provided to the algorithm. 

This setting is more challenging than Huber's contamination model, in that strong contamination allows the adversary to a) see the realization of the samples before choosing their contamination strategy (sometimes called an \emph{adaptive} adversary); and b) modify points (or, equivalently, add \emph{and} remove points), whereas a Huber adversary can only add points.\footnote{Though it is beyond the scope of this piece, understanding how these differences in capability affect the power of an adversary remain at the frontier of research\new{~\cite{BlancLMT22,BenDavidBKL23, BlancV25,LechnerBK25}}.}
In particular, this captures the setting where samples are drawn from a distribution which is $\eta$-close to $\cD$ in \emph{total variation distance}, thus expressing situations such as model misspecification.
Despite the power of the adversary in the strong contamination model, we will see that it is still possible to achieve compelling algorithmic results.

\subsection{Univariate Contamination-Robust Estimation}
With this contamination model in place, which algorithms are robust? And what error do they achieve?
We will start simple by focusing on Gaussian mean estimation in the univariate case (i.e., $d=1$), where the variance is known (and, without loss of generality, equal to 1).

A natural question is to ask whether the empirical mean is still suitable for this situation.
It is easy to see that it is \emph{not} contamination robust: consider an adversary who modifies a single sample to be extremely far away from the rest of the dataset.
This will shift the empirical mean by a large amount in the corresponding direction.
Thus, by corrupting only one datapoint (i.e., $\eta = 1/n$), the adversary can cause the error of the empirical mean to be arbitrarily large.

Fortunately, there \emph{are} contamination-robust estimators for the mean. 
The simplest example is the \emph{median}: let our estimator $\hat \mu$ be the $50$th percentile of the dataset. 
It is not hard to argue that the previous single-point contamination scheme will  not be able to shift the median of a Gaussian dataset that much, but the median also enjoys much stronger robustness properties\new{~\cite{Huber09}}.
Its (folklore) guarantees are formalized in the following statement.
\begin{proposition}[e.g., Corollary 1.15 of~\cite{DiakonikolasK22}]
    For any $\eta < 1/3$, let $X_1, \dots, X_n$ be an $\eta$-strong contamination of $n$ samples from $\normal(\mu, 1)$, for some $\mu \in \R$.
    If $\hat \mu = \operatorname{median}(X_1, \dots, X_n)$, we have that, with probability $\geq 95\%$, 
    \[|\hat \mu - \mu| \leq O\left(\frac{1}{\sqrt{n}} + \eta \right). \]
\end{proposition}
This proposition can be shown by reasoning about concentration of the empirical CDF (e.g., using the DKW inequality~\cite{DvoretzkyKW56}) of the uncontaminated points, and then, while accounting for the error introduced due to contamination, inverting the CDF of the Gaussian distribution. 

The rate enjoyed by the empirical median is the sum of two terms, a recurring pattern we will see for many robust estimators.
The first term is the standard rate from the non-contamination-robust setting ($1/\sqrt{n}$, sometimes called the \emph{parametric rate}), and we will refer to the excess term (in this case, $\eta$) as the \emph{cost of contamination}.\footnote{Throughout this article, we will de-emphasize focus on certain constant factors. For example, we will elide precise constant factors in the cost of contamination, though they are known in some cases (e.g., see~\cite{DiakonikolasKKLMS18} for analysis of the median). 
Similarly, the \emph{breakdown point} of estimators (the maximum value of $\eta$ that can tolerated) enjoys significant study in the literature\new{~\cite{Donoho82,ZhuJS20,ChenSS25}} but will not be a focus of our exposition.
Such constants may be important \new{in} practical considerations.}
It is not hard to show that this rate can not be improved -- optimality of the parametric rate is \new{textbook (see, e.g., Example 15.4 of~\cite{Wainwright19})}, and there exists an $\eta$-strong contamination of $\normal(\mu, 1)$ whose mean is at distance $\Omega(\eta)$, demonstrating optimality of the latter term. 

In short, the median completely resolves the problem of contamination-robust Gaussian mean estimation in the univariate setting. 
More generally, order statistics can be used as contamination-robust estimators for other univariate statistics -- for example, the interquartile range can be used to robustly estimate the variance of a Gaussian.

\subsection{Multivariate Contamination-Robust Estimation}
Given such strong results for univariate contamination-robust estimation, the natural question is whether they can be extended to the multivariate setting. 
Again, we will focus on Gaussian mean estimation, with covariance known to be the identity matrix $I$.
A natural first approach is to consider a multivariate generalization of the median.

One such generalization is Tukey's median~\cite{Tukey75}.
Informally speaking, Tukey's median is the ``deepest'' point in a dataset.
More formally, we let the \emph{Tukey depth} $\tau$ of a point $\theta \in \mathbb{R}^d$ with respect to a dataset $X$ be 
\[\tau(X, \theta) \triangleq \min_{\|v\|=1} \frac1n \sum_{i=1}^n \mathbbm{1}\{v^T X_i \leq v^T \theta\}.\]
For a point $\theta$, we consider all halfspaces that go through $\theta$ and pick the one that minimizes the fraction of datapoints it contains. 
The Tukey depth $\tau$ of $\theta$ is this minimum fraction of datapoints.
This is a natural measure of depth: no matter which direction you look in from $\theta$, you see at least $\tau$-fraction of the dataset.

The \emph{Tukey median} is then the point of maximum Tukey depth:
\[\hat \theta(X) = \arg \max_{\theta \in \mathbb{R}^d} \tau(X, \theta). \]
Observe that, in the univariate setting, the Tukey median reverts to the median.

The Tukey median enjoys strong guarantees on its robustness to contamination, summarized in the following theorem.
\begin{theorem}[Theorem 2.1 of~\cite{ChenGR18}, Theorem 3 of~\cite{ZhuJS22}]
\label{thm:tukey-median}
For any $\eta < 1/4$, let $X_1, \dots, X_n$ be an $\eta$-strong contamination of $n$ samples from $\normal(\mu, I)$, for some $\mu \in \R^d$.
    If $\hat \mu$ is the Tukey median of $X_1, \dots, X_n$, we have that, with probability $\geq 95\%$, 
    \[\|\hat \mu - \mu\|_2 \leq O\left(\sqrt{\frac{d}{n}} + \eta\right). \]
\end{theorem}
Once again, the rate is the sum of the non-contamination-robust parametric rate of $\sqrt{d/n}$ and the cost of contamination $\eta$.
Note that the linear dependence on $\eta$ is the exact same as in the univariate setting, so this estimator is again optimal.

While the Tukey median resolves the statistical problem, its computational aspects leave much to be desired.
Reflecting on the definition, it attempts to find a point which is central to the dataset in every univariate projection -- it is not clear how to compute such a point.
This intuition can be formalized: it is NP-hard to compute the Tukey median of a set of points~\cite{Bernholt06}.
Therefore, computing the Tukey median in even moderate dimensions is intractable.

Some alternatives one could consider include the coordinate-wise median, or the geometric median, the point which minimizes the sum of the $\ell_2$-distances to points in the dataset:\footnote{The geometric median can be defined similarly for a probability distribution, rather than a dataset.}
\[\arg \min_\nu \sum_{i=1}^n \|X_i - \nu\|_2. \]
While these quantities are all computationally tractable (see, e.g., ~\cite{CohenLMPS16} for an efficient algorithm to compute the geometric median), their cost of contamination is necessarily polynomial in the dimension. 
The following shows such a barrier for the geometric median, similar results hold for the coordinate-wise median and other variants.
\begin{proposition}[Proposition 2.1 of~\cite{LaiRV16}, \new{see also Lemma 1 of~\cite{PrasadBR20} for a more general statement}]
    Let $\mu \in \mathbb{R}^d$ and $\Sigma \in \mathbb{R}^{d\times d}$ be a diagonal covariance matrix with $0$ in the first coordinate and $1$ in all other coordinates.
    There exists a distribution which is $\eta$-close to $\normal(\mu, \Sigma)$ in total variation distance with geometric median $\hat \mu$ such that 
    \[\|\hat \mu - \mu\|_2 \geq \Omega\left(\eta\sqrt{d}\right).\]
\end{proposition}

This implies that, even in the infinite data limit (i.e., $n \rightarrow \infty$), the $\ell_2$ error in the estimate of the mean (i.e., the cost of contamination) is $\Omega(\sqrt{d})$.
There are many different algorithms that achieve a similar cost of contamination: arguably the simplest method involves discarding any sample points that are sufficiently far from a majority of the dataset, and taking the empirical mean of the remainder.
Contrast this dimension-dependent guarantee with the Tukey median, whose cost of contamination depends only on the fraction of contamination and not the dimension (Theorem~\ref{thm:tukey-median}).

\paragraph{A contamination dilemma} To summarize, all methods we have seen so far (and indeed, all methods prior to 2016) for contamination-robust mean estimation suffered from one of the following two deficiencies:
\begin{enumerate}
    \item The running time of the method scales exponentially in the dimension; or
    \item The cost of contamination incurred by the method scales polynomially in the dimension.
\end{enumerate}
Either of these deficiencies renders an approach impractical, even for settings of moderate dimensionality.
Despite decades of study, this stood as a longstanding roadblock to realizing contamination robustness for estimation tasks on multivariate data.
In a 1997 retrospective~\cite{Huber97}, pioneer of the field Peter Huber lamented this state of affairs.
\begin{quote}
    ``It is one thing to design a theoretical algorithm whose purpose is to prove [large fractions of
corruptions can be tolerated] and quite another thing to design a practical version that can be
used not merely on small, but also on medium sized regression problems, with a 2000 by 50 matrix
or so. This last requirement would seem to exclude all of the recently proposed [techniques].''
\end{quote}

\subsection{Efficient Multivariate Contamination-Robust Estimation}

In 2016, concurrent works on contamination-robust mean estimation by Diakonikolas, Kamath, Kane, Li, Moitra, and Stewart (DKKLMS)~\cite{DiakonikolasKKLMS16} and Lai, Rao, and Vempala~\cite{LaiRV16} resolved this tension. 
DKKLMS provided the first computationally efficient algorithm for contamination-robust mean estimation featuring a dimension-independent cost of contamination.\footnote{Lai, Rao, and Vempala~\cite{LaiRV16} give a computationally efficient algorithm where the cost of contamination depends mildly on the dimension ($O(\eta \sqrt{\log d})$), rather than the dimension independent ($O(\eta \sqrt{\log 1/\eta})$) result of DKKLMS.}

\begin{theorem}[Theorem 4.22 of~\cite{DiakonikolasKKLMS16}]
\label{thm:robust-mean}
    For any $\eta$ less than an absolute constant, let $X_1, \dots, X_n$ be an $\eta$-strong contamination of $n$ samples from $\normal(\mu, I)$, for some $\mu \in \R^d$.
    There exists a polynomial-time estimator $\hat \mu(X_1, \dots, X_n)$ such that, with probability $\geq 95\%$, 
    \[\|\hat \mu - \mu\|_2 \leq O\left(\sqrt{\frac{d}{n}} + \eta\sqrt{\log 1/\eta}\right). \]
\end{theorem}

 Observe that a) this method is computable in polynomial time, and b) the cost of contamination is only $\eta \sqrt{\log 1/\eta}$, independent of the dimension $d$.\footnote{There is evidence that this extra $\sqrt{\log 1/\eta}$ factor is unavoidable for computationally efficient estimators~\cite{DiakonikolasKS17}.}

Broadly speaking, computationally efficient contamination-robust estimators can be designed by first reasoning about properties of the dataset generated from a distribution, and then employing them to algorithmically limit the impact of contamination.
We will explore contamination-robust Gaussian mean estimation through the lens of a notion of \emph{stability} (following the presentation of~\cite{DiakonikolasK22}, \new{note also the closely related concept of \emph{resilience}~\cite{SteinhardtCV18}}).
We will first describe how Gaussian data enjoys stability of its empirical mean and covariance.
\new{This will allow us to expose the core solution concepts that underlie the resulting algorithms, and set things up to subsequently discuss connections with other forms of robustness.}
We then show how to use \new{these properties} to design an inefficient contamination-robust mean estimator. 
Finally, \new{we discuss a simple spectral method based on these ideas}, that can efficiently compute a contamination-robust estimate of the mean.

It is well understood that a dataset generated according to a Gaussian distribution will have an empirical mean and covariance concentrated around the true mean and covariance with high probability (see, e.g.,~\cite{Vershynin18}).
In fact, a stronger property holds: all \emph{sufficiently large} subsets of the dataset will enjoy such concentration. 
Indeed, it is possible to show that, with high probability, a set of $O(d/\gamma^2)$ samples from $\normal(\mu, I)$ satisfies the following notion of stability with respect to $\mu$.
\begin{definition}[Definition 2.1 of~\cite{DiakonikolasK22}]
\label{def:stability}
    Fix $\gamma \leq 1/2$.
    Let $X_1, \dots X_n \in \mathbb{R}^d$ be a dataset, and for a set $S \subseteq [n]$, let  $\mu_S = \frac{1}{|S|} \sum_{i \in S} X_i$ be its empirical mean and $\Sigma_S = \frac{1}{|S|} \sum_{i \in S} (X_i - \mu_S) (X_i - \mu_S)^T$ be its empirical covariance.
    $X_1, \dots, X_n$ is $(\gamma, O(\gamma \sqrt{\log 1/\gamma}))$-stable with respect to a vector $\mu$ if the following holds for all subsets $S \subseteq [n]$ such that $|S| \geq (1-\gamma)n$:
    \begin{itemize}
        \item $\|\mu_S - \mu\|_2 \leq O(\gamma \sqrt{\log 1/\gamma})$;
        \item $\|\Sigma_S - I\|_2 \leq O(\gamma \log 1/\gamma)$.
    \end{itemize}
\end{definition}
Proving that a Gaussian dataset satisfies this definition appeals to arguments involving the tail bounds of the Gaussian.
Indeed, qualitatively similar results can be shown for other ``nice'' distributions (i.e., satisfying weaker moment bounds) which lead to contamination-robust estimators for those settings.
Observe that the first condition of this stability property immediately implies that the empirical mean is robust to \emph{subtractive} contaminations.

It turns out that stability can also be used to reason about properties of \emph{general} contaminations of stable datasets.
\begin{lemma}[Lemma 2.6 of~\cite{DiakonikolasK22}]
\label{lem:large-ev-stab}
Let $X_1, \dots, X_n$ be a $\gamma$-strong contamination of a $(\gamma, O(\gamma \sqrt{\log 1/\gamma}))$-stable dataset with respect to a vector $\mu$.
Let $\mu'$ and $\Sigma'$ be the empirical mean and covariance of this dataset.
If  $\|\Sigma'\|_2 \leq 1 + \gamma \log 1/\gamma$, then $\|\mu' - \mu\|_2 \leq O(\gamma \sqrt{\log 1/\gamma})$.
\end{lemma}
Essentially, this lemma says that if a \emph{contamination} of a stable dataset does not have large \new{variance in any direction (i.e., the eigenvalues of its empirical covariance matrix are not too large)}, then its empirical mean will be close to the true mean.

Lemma~\ref{lem:large-ev-stab} immediately gives a (computationally inefficient) algorithm for contamination-robust mean estimation.
For every sufficiently large subset of a contaminated dataset (which is itself a contaminated dataset with larger contamination parameter $\eta$), compute the top eigenvalue of its empirical covariance matrix: if it is less than $1 + O(\eta \log 1/\eta)$, output its empirical mean.
A subset with this property is guaranteed to exist by the definition of stability.

Though we will not use it immediately, we introduce a related and more general notion of a \emph{spectral center}, which is commonly seen across the literature in robust estimation.
\begin{definition}[Definition 5.1 of \cite{HopkinsLZ20}]
\label{def:spectral-center}
    A point $\nu \in \R^d$ is a $(\gamma, \lambda)$-spectral center of $X_1, \dots, X_n \in \R^d$ if
    \[\min_{w \in \mathcal{W}_{n, \gamma}} \left\|\sum_{i=1}^n w_i (X_i - \nu)(X_i - \nu)^T \right\|_2 \leq \lambda  ,\]
    where $\mathcal{W}_{n, \gamma} = \left\{w \in \Delta^n\ :\ \|w\|_\infty \leq \frac{1}{(1-\gamma)n}\right\}$.\footnote{We use $\Delta^n$ to denote the probability simplex over $[n]$.}
\end{definition}
The approach described above demonstrates that, if $w$ is uniform over a large subset of the points and $\nu$ is their empirical mean $\mu'$, then $\mu'$ is a spectral center. 
However, this definition also allows for non-uniform  weightings, which can be useful for methods that consider soft removal of points. 
For the time being, we will return to the special case of finding a subset of our contamination of a stable dataset that has bounded top eigenvalue.

As an aside, we note that a spectral center satisfies a centrality condition which holds for all directions: an equivalent interpretation is that, for all unit vectors $v \in \R^d$, the variance of the dataset $v \cdot X_1, \dots, v \cdot X_n$ is at most $\lambda$.
The fact that this property holds for all directions is reminiscent of the Tukey median, and in contrast to the coordinate-wise or geometric median.
It thus seems important for achieving a dimension-independent cost of contamination.

Lemma~\ref{lem:large-ev-stab} gives us a hint towards designing an \emph{efficient} algorithm for contamination-robust mean estimation.
If the top eigenvalue of the empirical covariance matrix of a contaminated dataset is small, we can output the dataset's empirical mean. 
This begs the question: what if the top eigenvalue is \emph{large}? 
This implies that, in the one-dimensional projection of the corresponding eigenvector $v$, the variance is larger than it ought to be if the data were truly Gaussian. 
Thus, we can project onto the top eigenvector $v$, and attempt to find and remove outliers in this simpler one-dimensional setting.

Once we focus on this particular one-dimensional projection (which, informally, ``looks off'' due to its larger-than-expected variance), there are many ways we can proceed. 
Our overarching goal is to remove contaminated points, or at least reduce their influence, in comparison to the uncontaminated portion of the dataset.
One way to do this is to note that the uncontaminated dataset follows a Gaussian distribution, and thus obeys the corresponding Gaussian tail bounds.
However, since the variance in this univariate projection $v$ is larger than it ought to be, there must exist a threshold $L$ such that the number of points $x$ where $|v \cdot (x - \mu')| \geq L$ is larger than these tail bounds prescribe (where $\mu'$ is the empirical mean of the dataset).
Removing all such points will remove more contaminated points than uncontaminated points, and serve as a measure of progress, repeating until all contaminated points are removed, or the top eigenvalue of the empirical covariance matrix is otherwise small.
There are variants of this strategy that one can employ after projecting in the high-variance direction $v$, including randomized thresholding, independent and randomized datapoint removal, and deterministic reweighting of datapoints.

Putting it all together, an algorithm for contamination-robust multivariate Gaussian mean estimation is described in Figure~\ref{alg:robust-mean}.

\begin{figure}
\caption{Algorithm for contamination-robust mean estimation}\label{alg:robust-mean}
\begin{algorithmic}[1]
 \renewcommand{\algorithmicrequire}{\textbf{Input:}}
 \renewcommand{\algorithmicensure}{\textbf{Output:}}
\Require{Dataset $X = \{X_1, \dots, X_n\} \in \R^{n \times d}$, contamination fraction $\eta \in [0,1]$}
 \Ensure{Mean estimate $\hat \mu \in \R^d$}
 \Procedure{CRMeanEstimation}{$X, \eta$}
 \State $\hat \mu \gets \frac{1}{n}\sum_{i=1}^n X_i$
 \State $\hat \Sigma \gets \frac{1}{n}\sum_{i=1}^n (X_i - \hat \mu)(X_i - \hat \mu)^T$
 \State $(\lambda, v) \gets$ the top eigenvalue/eigenvector pair of $\hat \Sigma$
 \If{$\lambda \leq 1 + O(\eta \log 1/\eta)$} 
 \State \Return{$\hat \mu$}
 \Else 
 \State Identify a threshold $L$ such that $\frac{1}{n} \sum_{i=1}^n\mathbbm{1}\{|v\cdot (X_i - \mu)| \geq L\}$ is much greater than $\Pr_{Z\sim \normal(\mu, I)}[|v\cdot (Z - \mu)| \geq L]$
 \State $Y \gets \{X_i\ :\ |v\cdot (X_i - \mu)| \leq L\}$
 \State \Return{\textsc{CRMeanEstimation}$(Y, \eta)$}
 \EndIf
 \EndProcedure
\end{algorithmic}
\end{figure}

\subsection{Beyond Gaussian Mean Estimation}
Thus far, our focus has been on contamination-robust mean estimation of multivariate Gaussians with identity covariance. 
However, these ideas are not limited to this specific case, and can be extended to many other settings.

For example, the assumption of identity covariance is quite restrictive: DKKLMS~\cite{DiakonikolasKKLMS16} further show that an arbitrary Gaussian distribution can be robustly estimated in total variation distance.
\begin{theorem}
    For any $\eta$ less than an absolute constant, let $X_1, \dots, X_n$ be an $\eta$-strong contamination of $n$ samples from $\normal(\mu, \Sigma)$, for some $\mu \in \R^d, \Sigma \in \R^{d\times d}$.
    There exists polynomial-time estimators $\hat \mu(X_1, \dots, X_n)$ and $\hat \Sigma(X_1, \dots, X_n)$ such that, with probability $\geq 95\%$, 
    \[\dtv(\normal(\mu, \Sigma), \normal(\hat \mu,  \hat\Sigma)) \leq \tilde O\left(\sqrt{\frac{d^2}{n}} + \eta\log 1/\eta\right). \]
\end{theorem}
Since contamination-robust mean estimation involves inspecting the second moment of the data for deviations, naturally, contamination-robust estimation of the covariance matrix requires inspecting the fourth moment. 
This case proves to be more challenging since we do not have explicit knowledge of the fourth moment -- nonetheless, Gaussians possess enough structure that we can upper bound the fourth moment in terms of a known function of the second moment, which allows us to start from a coarse estimate and iteratively improve it.

Gaussianity is another strong assumption. While the algorithms described above also work for sub-Gaussian distributions,\footnote{Recall the definition of sub-Gaussian random variables in Definition~\ref{def:sub-gaussian}.} there exist variants that work for distributions with only a bound on their variance.
\begin{theorem}[\cite{DiakonikolasKKLMS17,SteinhardtCV18}]
\label{thm:second-moment-robust}
    For any $\eta$ less than an absolute constant, let $X_1, \dots, X_n$ be an $\eta$-strong contamination of $n$ samples from a distribution $\cD$ with mean $\mu \in \R^d$ and covariance $\Sigma \preceq I$.
    There exists a polynomial-time estimator $\hat \mu(X_1, \dots, X_n)$ such that, with probability $\geq 95\%$, 
    \[ \|\hat \mu - \mu \| \leq O\left(\sqrt{\frac{d}{n}} + \sqrt{\eta}\right). \]
\end{theorem}
Observe that the cost of contamination increases from $\eta \log 1/\eta$ to $\sqrt{\eta}$ -- this is inherent, due to the weaker assumptions on the moments of the distribution.

The applicability of these techniques is not limited to estimation tasks, and can also be used for supervised learning: Klivans, Long, and Servedio~\cite{KlivansLS09} previously employed similar ideas for robust learning of halfspaces.

Our discussion here only scratches the surface of recent work on algorithmic contamination-robust statistics -- see~\cite{DiakonikolasK22} for a more thorough coverage of the topic.

\section{Sub-Gaussian Rates for Heavy-Tailed Mean Estimation}
\label{sec:subgaussian}
We now turn our attention to another form of robustness.
Up to this point, we have focused on mean estimation of \emph{Gaussian} distributions.
Conveniently,  Gaussians have very \emph{light tails}: samples from a Gaussian are highly unlikely to be extremely far away from the mean, due to (super-)exponentially decaying tails.
This property implies that many procedures applied to Gaussian data automatically inherit very sharp concentration, including, most germane to our discussion, the empirical mean.
On the other hand, data in the wild may be far more ill-behaved.
Specifically, data may come from distributions with \emph{heavy tails}, allowing ``outlier'' points extremely far from the mean to arise with much higher probability. 
Such outliers introduce significantly more error for statistics like the empirical mean.
Can we achieve error rates comparable to the Gaussian case, robust to outliers caused by heavy-tailed data?

To formalize the above discussion, we revisit our original problem formulation. 
We have focused thus far on mean estimation with high constant probability of success, e.g., $\geq 95\%$. 
We will now aim for arbitrarily high probability of success, i.e., $\geq 1 - \beta$ for some $\beta > 0$.
If we analyze the empirical mean for Gaussian data, we find that it enjoys a very mild  dependence on $1/\beta$.
In particular, the required amount of data to achieve a particular accuracy guarantee with high probability is inflated only by a factor of $\log 1/\beta$, which can be shown to be optimal. 
\begin{proposition}
\label{prop:sub-gaussian}
    Let $X_1, \dots, X_n$ be $n$ samples from $\normal(\mu, 1)$, for some $\mu \in \R$.
    If $\hat \mu = \frac1n \sum_{i=1}^n X_i$, we have that, with probability $\geq 1-\beta$, 
    \[|\hat \mu - \mu| \leq O\left(\sqrt{\frac{\log 1/\beta}{n}}\right). \]
\end{proposition}
This is easy to prove: since $\hat \mu$ is the average of $n$ samples from $\normal(\mu, 1)$, it is distributed as $\normal(\mu, 1/n)$.
The proposition follows from a Gaussian tail bound.

We recall the following relaxation of Gaussian distributions, namely \emph{sub-Gaussian} distributions.
Such distributions are defined by having tails that are at least as light as those of a Gaussian distribution.
\begin{definition}\label{def:sub-gaussian}
    A random variable $X \in \R$ is sub-Gaussian with variance proxy $K$ if $\E\left[e^{\left(X - \E[X]\right)t}\right] \leq \exp\left(\frac{K^2t^2}{2}\right)$ for all $t$, where $K$ is a positive constant.

    A random vector $X \in \R^d$ is sub-Gaussian with variance proxy $K$ if the random variable $\langle v, X\rangle$ is sub-Gaussian with variance proxy $K$ for all unit vectors $v \in \R^d$.
\end{definition}

Proposition~\ref{prop:sub-gaussian} holds more generally, for mean estimation of any sub-Gaussian distribution (with variance proxy 1), and this rate is often referred to as the \emph{sub-Gaussian} rate.

What if the underlying distribution has heavy tails? 
This is possible if, for example, the distribution has only a bound on its variance.
%Of course, sub-Gaussianity is a strong assumption for a distribution -- the question is whether a similar rate can be achieved even with weaker assumptions on its moments.
%The most common such assumption is bounded variance. 
Unfortunately, formalizing the deficiency described above, the empirical mean has an exponentially worse dependence on $1/\beta$ in this case.
\begin{proposition}
    Let $X_1, \dots, X_n$ be $n$ samples from a distribution $\cD$ with mean $\mu \in \R$ and variance $\sigma^2 \leq 1$.
    If $\hat \mu = \frac1n \sum_{i=1}^n X_i$, we have that, with probability $\geq 1-\beta$,
    \[ |\hat \mu - \mu | \leq O\left(\sqrt{\frac{1}{\beta n}}\right). \]
    \new{Furthermore, there exists such a distribution $\cD$ such that, with probability $\geq 1 - \beta$,
    \[|\hat \mu - \mu| \geq \Omega \left(\sqrt{\frac{1}{\beta n}}\right).\]}
\end{proposition}
\new{The upper bound in this proposition is again shown via a tail bound on the averaged distribution $\hat \mu$, which has variance at most $1/n$.
The result follows from Chebyshev's inequality.
This analysis is tight for the case when we have bounds on only the second moments~\cite{Catoni12}.}
%Intuitively, this degraded accuracy is due to the fact that tail events occur with a relatively high probability, and can introduce significant error to the empirical mean.
%Thus, in the Statistics community, robustness is sometimes used to refer to minimizing error even when the data is heavy tailed.

The natural question: is it possible to achieve the sub-Gaussian rate for mean estimation, even when the distribution only has bounded variance?
This would allow us to get an accurate estimate of the mean with very high probability, robust to potentially heavy tails of the underlying distribution.

The \emph{median of means} paradigm solves this problem: split the $n$ datapoints into $k = \Omega(\log 1/\beta)$ batches, compute their respective means (sometimes called \emph{bucket means}), and output the median of the results.  
\begin{proposition}[\new{e.g., Exercise 2.2.9 of~\cite{Vershynin18}}]
    Let $X_1, \dots, X_n$ be $n$ samples from a distribution $\cD$ with mean $\mu \in \R$ and variance $\sigma^2 \leq 1$.
    Let $Y_1, \dots, Y_k$ be the bucket means after partitioning $X_1, \dots, X_n$ into $k = \Theta(\log 1/\beta)$ parts.
    If $\hat \mu = \textsc{Median}(Y_1, \dots, Y_k)$, we have that, with probability $\geq 1-\beta$,
    \[ |\hat \mu - \mu | \leq O\left(\sqrt{\frac{\log 1/\beta}{ n}}\right). \]
\end{proposition}
To sketch the analysis: by Chebyshev's inequality, each bucket mean $Y_i$ will be within distance $O(\sqrt{k/n})$ of the true mean $\mu$ with high constant probability (say, $\geq 90\%$), and thus with very high probability ($1 - e^{-\Omega(k)}$), most bucket means will be in an interval of width $O(\sqrt{k/n})$ of the true mean.
If this is the case, taking the median of the bucket means will produce a point in this interval.
Taking $k = \Theta(\log 1/\beta)$ gives the proposition.

This analysis suggests another (more relaxed) algorithm that achieves the same result  for the univariate setting (up to constant factors): output \emph{any} point which is within distance $O(\sqrt{k/n})$ of the majority of the bucket means.

Mirroring the case of contamination-robustness in Section~\ref{sec:robust}, in the univariate case, issues with the empirical mean are relatively easy to resolve via appeal to robust statistics like the median.
Unfortunately, the similarities do not end there: we again run into issues when we try to generalize and extend to the multivariate setting. 

\subsection{Sub-Gaussian Rates for Multivariate Data}
In $d$-dimensional settings, the sub-Gaussian rate (achieved by the empirical mean for sub-Gaussian distributions) is
\begin{equation} \label{eq:subgaussian}
  \left\|\frac{1}{n}\sum_{i=1}^n X_i - \mu \right\|_2 \leq O\left(\sqrt{\frac{d}{n}} + \sqrt{\frac{\log 1/\beta}{n}}\right).  
\end{equation}

Compare this with the rate of the empirical mean for distributions with only bounded second moment:
\begin{equation}\label{eq:heavy-tailed}
  \left\|\frac{1}{n}\sum_{i=1}^n X_i - \mu \right\|_2 \leq O\left(\sqrt{\frac{d}{\beta n}}\right).  
\end{equation}
Once again, we see that the dependence on $\beta$ degrades from logarithmic to polynomial.
But, additionally, we also see that the ideal sub-Gaussian rate of \eqref{eq:subgaussian} \emph{decouples} the dependence on $d$ and $1/\beta$ -- the $\log 1/\beta$ thus effectively serves as an \emph{additive} overhead in the amount of data $n$ required.
The same is not true for the rate \eqref{eq:heavy-tailed} of the empirical mean for heavy-tailed distributions, where it incurs a \emph{multiplicative} overhead.

So how do we achieve the sub-Gaussian rate for heavy-tailed data, even in the multivariate setting?
We will try to extend the median of means paradigm.
Once again, we will compute the bucket means of several subsets of the data, and combine them using some sort of median.
We can start by trying the alternative to the standard median proposed above: output any point which is close to the majority of the bucket means in $\ell_2$ norm (sometimes, such a point is called a \emph{simple median}).
The argument is similar to before: each bucket mean will be within $\ell_2$-distance $O(\sqrt{d(k/n)})$ of the true mean $\mu$ with high constant probability, and therefore with very high probability ($1-e^{-\Omega(k)}$), most bucket means will be in an $\ell_2$ ball of comparable radius.
Taking $k = \Theta(\log 1/\beta)$ gives the following guarantee.
\begin{proposition}
    Let $X_1, \dots, X_n$ be $n$ samples from a distribution $\cD$ with mean $\mu \in \R^d$ and covariance $\Sigma \preceq I$.
    Let $Y_1, \dots, Y_k$ be the bucket means after partitioning $X_1, \dots, X_n$ into $k = \Theta(\log 1/\beta)$ parts. 
    If $\hat \mu = \textsc{SimpleMedian}(Y_1, \dots, Y_k)$, we have that, with probability $\geq 1-\beta$,
    \[ |\hat \mu - \mu | \leq O\left(\sqrt{\frac{d \log 1/\beta}{ n}}\right). \]
\end{proposition}
We see that, while we incur the desired logarithmic dependence on $1/\beta$, in contrast to \eqref{eq:subgaussian}, it \emph{multiplies} the dimension $d$. 
Other simple strategies for aggregating the bucket means, such as the geometric or coordinate-wise median, suffer from the same deficiency.

In 2017, Lugosi and Mendelson~\cite{LugosiM17, LugosiM19a} proposed a variant of what is now called a \emph{combinatorial center}.
This can be seen as a refinement of the simple median: it seeks to find a point which is close to a majority of the bucket means \emph{in every univariate projection}.
More formally, we have the following definition.
\begin{definition}[Definition 5.2 of \cite{HopkinsLZ20}]
\label{def:combinatorial-center}
    A point $\nu \in \R^d$ is a $(\gamma, \lambda)$-combinatorial center of $Y_1, \dots, Y_k \in \R^d$ if for all unit vectors $v \in \R^d$,
    \[\sum_{i=1}^k \mathbbm{1}\left\{\langle Y_i - \nu, v\rangle \geq \sqrt{\lambda} \right\} \leq \gamma k.\]
\end{definition}

\new{The combinatorial center is an essential concept in many recent algorithms achieving sub-Gaussian rates.}
The following key lemma, showing that the mean $\mu$ is a combinatorial center of the bucket means, is central to the approach introduced by Lugosi and Mendelson~\cite{LugosiM19a}.
\begin{lemma}\label{lem:comb-cent}
    Let $X_1, \dots, X_n$ be $n$ samples from a distribution $\cD$ with mean $\mu \in \R^d$ and covariance $\Sigma \preceq I$. 
    Let $Y_1, \dots, Y_k$ be the bucket means after partitioning $X_1, \dots, X_n$ into $k$ parts. 
    With probability at least $1 - e^{-\Omega(k)}$, $\mu$ is a $(0.01, r_k^2)$-combinatorial center of $Y_1, \dots, Y_k$, where

    \[r_k = O\left(\sqrt{\frac{d}{n}}+ \sqrt{\frac{k}{n}} \right).\]
\end{lemma}

%Lugosi and Mendelson~\cite{LugosiM19a} showed that a) the mean of a distribution will be a combinatorial center of the bucket means with high probability, and b) a combinatorial center of the bucket means achieves the sub-Gaussian rate. 
%Putting these together, one can show the following theorem:

Setting $k = \Omega(\log 1/\beta)$ and reasoning that any two combinatorial centers must be close gives the following result.

\begin{theorem}[\cite{LugosiM19a}]
    Let $X_1, \dots, X_n$ be $n$ samples from a distribution $\cD$ with mean $\mu \in \R^d$ and covariance $\Sigma \preceq I$. 
    Let $Y_1, \dots, Y_k$ be the bucket means after partitioning $X_1, \dots, X_n$ into $k = \Theta(\log 1/\beta)$ parts. 
    There exists an estimator $\hat \mu(Y_1, \dots, Y_k)$ such that, with probability $\geq 1 - \beta$, 
    \[\|\hat \mu - \mu \|_2 \leq O\left(\sqrt{\frac{d}{n}} + \sqrt{\frac{\log 1/\beta}{n}}\right). \]
\end{theorem}
In other words, using the combinatorial center in the median-of-means paradigm, we can achieve the sub-Gaussian rate for mean estimation with heavy-tailed data. 
The main caveat is once again computational in nature.
Similar to the Tukey median, a combinatorial center is central to the dataset in \emph{every one-dimensional projection}, which creates computational challenges: it is not clear how to efficiently locate a combinatorial center, or even \emph{certify} that a point is one (that is, given a point purported to be a combinatorial center, verify that it indeed satisfies the definition). 
In fact, it can be viewed as a slight relaxation of the Tukey median: in cases where a point of high Tukey depth exists, a combinatorial center is close to a Tukey median in every one-dimensional projection.
Lugosi and Mendelson focused entirely on statistical properties of the combinatorial center, and left open whether such an object could be efficiently computed.

\paragraph{A heavy-tailed dilemma}
 To summarize, we face barriers reminiscent to those we described for robustness. Prior to 2018, all existing methods for mean estimation of heavy-tailed distributions suffered at least one of the following deficiencies:
\begin{enumerate}
    \item The running time of the method scales exponentially in the dimension; or
    \item The dependence on the (inverse of the) failure probability $\beta$ is greater than in the sub-Gaussian rate, either linear rather than logarithmic, or being multiplied by a dimension-dependent factor.
\end{enumerate}

\subsection{Efficient Sub-Gaussian Rates for Multivariate Data}
In 2018, Hopkins~\cite{Hopkins18, Hopkins20} managed to bypass these barriers, giving an efficient algorithm for computing a combinatorial center, and thus the first computationally efficient algorithm for mean estimation of heavy-tailed distributions with sub-Gaussian rates.
\new{We will start by describing an inefficient approach to find a combinatorial center, and then discuss ideas that allow us to make it efficient.}

Let's start with a simpler problem: suppose someone gave you a point $\nu \in \R^d$, which they claimed was a $(\gamma, \lambda)$-combinatorial center of a dataset $Y_1, \dots, Y_k$.
How would you \emph{certify} that this was the case? 
This can be done (inefficiently) using the following quadratic program \eqref{prog:qp}.
\begin{align}
    \max_{v \in \R^d, b \in \R^k} &\sum_{i=1}^k b_i  \text{ subject to} \label{prog:qp} \tag{QP-CC} \\
    b_i &\in \{0, 1\} \ &\forall i \in [k] \nonumber \\
    \|v\|_2^2 &= 1 \nonumber \\
    b_i \sqrt{\lambda} &\leq \langle Y_i - \nu, b_i v \rangle\ &\forall i \in [k] \nonumber 
\end{align}
In particular, the objective function is at most $\gamma k$ if and only if $\nu$ is a $(\gamma, \lambda)$-combinatorial center.

To interpret this quadratic program, the first constraint enforces that the $b_i$'s are indicators, and the second ensures that $v$ is a unit vector.
The most important constraint is the third: if $b_i = 1$, this ``counts'' (via the objective function) that $Y_i$ is distant from $\nu$ in the direction $v$ (otherwise, if $b_i = 0$, the condition is vacuously satisfied).
The optimization problem tries to find the direction $v$ (and the corresponding count) that maximizes the number of points that exceed this distance condition. 
If the worst case over all directions $v$ is at most $\gamma k$, then we certify that $\nu$ satisfies the definition of a $(\gamma, \lambda)$-combinatorial center.

\new{Given this procedure that \emph{certifies} that a point is a combinatorial center, how do we \emph{find} a combinatorial center?
If we are not concerned with computation, a straightforward approach is to discretize the space and try every point $\nu$ as a candidate combinatorial center in the quadratic program~\eqref{prog:qp}.
However, we will instead employ an oracle-efficient descent-based approach introduced in 2019 by Cherapanamjeri, Flammarion, and Bartlett~\cite{CherapanamjeriFB19}.
This has the advantage that, given an oracle that solves the quadratic program~\eqref{prog:qp}, one can find a combinatorial center with only $O(\log d)$ calls to the oracle.
This leaves us with only one source of computational intractability (solving the quadratic program) rather than introducing a new one.

Suppose we tried to certify a point $\nu$ which is \emph{not} a combinatorial center. 
What would a solution to the quadratic program~\eqref{prog:qp} look like?
In addition to returning the indicator $b_i$'s (telling us which points are ``far''), it produces a vector $v$, pointing in the direction of these far points.
Intuitively, taking a step in this direction to $\nu + \zeta v$ (where $\zeta$ is some judiciously chosen step size) should give us a point which is ``more central'' to the dataset. 
We can iterate this process on $\nu + \zeta v$, repeatedly obtaining a new direction to step in and eventually finding and certifying a point $\hat \mu$ to be a combinatorial center.

Now that we have an inefficient algorithm, we turn our attention to making the procedure efficient. 
At the heart of this algorithm, the intractability arises due to the quadratic program for certification~\eqref{prog:qp}.
The canonical way to make an optimization problem based on quadratic programming efficient is by instead considering its semidefinite programming (SDP) relaxation. 
}
%Instead, Hopkins considers a semidefinite programming (SDP) relaxation of this optimization problem.
Let $v \in \R^d, V \in \R^{d \times d}, b \in \R^k, B \in \R^{k \times k}, W \in \R^{k \times d}$ be the optimization variables. 
\new{Hopkins~\cite{Hopkins20}  introduced the following optimization problem.}
\begin{align}
    \max &\sum_{i=1}^k b_i  \text{ subject to} \label{prog:sdp} \tag{SDP-CC} \\
    B_{ii} &= b_i \ &\forall i \in [k] \nonumber \\
    \operatorname{Tr}(V) &= 1 \nonumber \\
    b_i \sqrt{\lambda} &\leq \langle Y_i - \nu, W_i \rangle\ &\forall i \in [k] 
 \nonumber \\
    &\begin{bmatrix}
1 & b^T & v^T\\
b & B & W \\
v & W^T & V 
\end{bmatrix} \succeq 0 \nonumber 
\end{align}
To see this is a relaxation of the quadratic program, consider a solution $v, b$ to the quadratic program~\eqref{prog:qp}, and let $V = vv^T$, $B = bb^T$, and $W = bv^T$.
The fact that $v$ and $b$ satisfy the constraints of the quadratic program~\eqref{prog:qp} imply they satisfy the corresponding constraints of the SDP~\eqref{prog:sdp}, and since the matrix in the final constraint can be written as $\left[\begin{smallmatrix}
  1 \\
  b \\
  v
\end{smallmatrix}\right]\left[\begin{smallmatrix}
  1 & b^T & v^T
\end{smallmatrix}\right]$, it is satisfied as well.
Importantly, since the relaxed program~\eqref{prog:sdp} is an SDP, we can efficiently find a solution to~\eqref{prog:sdp}. 

We introduce a definition for a combinatorial center that can also be certified by~\eqref{prog:sdp}.

\begin{definition}
    A point $\nu$ is a certifiable $(\gamma, \lambda)$-combinatorial center of $Y_1, \dots, Y_k \in \R^d$ if the value of \eqref{prog:sdp} is at most $\gamma k$.
\end{definition}

Observe that, since~\eqref{prog:sdp} is a relaxation of~\eqref{prog:qp} (and thus, its objective function is larger), every certifiable $(\gamma, \lambda)$-combinatorial center is also a (vanilla) $(\gamma, \lambda)$-combinatorial center.
However, the reverse is not necessarily true: a point may be a combinatorial center, but the objective function of~\eqref{prog:sdp} may be large, and thus, we would be unable to certify its centrality.
Remarkably, the following lemma shows that this is not the case for the true mean $\mu$, which is certifiably central with high probability.
\begin{lemma}[Lemma 2.8 of~\cite{Hopkins20}]
Let $X_1, \dots, X_n$ be $n$ samples from a distribution $\cD$ with mean $\mu \in \R^d$ and covariance $\Sigma \preceq I$. 
    Let $Y_1, \dots, Y_k$ be the bucket means after partitioning $X_1, \dots, X_n$ into $k = \Theta(\log 1/\beta)$ parts. 
    Letting $r_k = \sqrt{\frac{d}{n}} + \sqrt{\frac{\log 1/\beta}{n}}$, we have that $\mu$ is a certifiable $(0.01, O(r_k^2))$-combinatorial center of $Y_1, \dots, Y_k$ with probability $\geq 1- \beta$.
\end{lemma}

\new{
This establishes how to efficiently solve the certification problem.
Hopkins~\cite{Hopkins20} goes on to use the powerful Sum-of-Squares (SoS) proofs-to-algorithms framework to \emph{find} a (certifiable) combinatorial center.\footnote{\new{See the textbook of Barak and Steurer for an introduction to SoS~\cite{BarakS16}.}}
This technique has enjoyed tremendous success in developing efficient algorithms for a range of problems within and beyond the realm of robustness~\cite{RaghavendraSS18}.
However, SoS-based algorithms can be technical and require significant background.

It turns out the aforementioned descent-based approach of Cherapanamjeri, Flammarion, and Bartlett~\cite{CherapanamjeriFB19} still works even on the relaxed SDP~\eqref{prog:sdp}.
Indeed, while initial works required a non-trivial rounding scheme to extract a direction from the matrix $V$, it was subsequently shown by Hopkins, Kamath, and Majid~\cite{HopkinsKM22} that one can use the direction $v$ in the SDP without further modification, despite the loss of the interpretation in the quadratic program described above.
}

By putting these pieces together, one can conclude a computationally efficient algorithm for mean estimation with sub-Gaussian rates, robust to potential outliers caused by heavy-tailed data.
\begin{theorem}[\cite{Hopkins20,CherapanamjeriFB19}]
    Let $X_1, \dots, X_n$ be $n$ samples from a distribution $\cD$ with mean $\mu \in \R^d$ and covariance $\Sigma \preceq I$. 
    There exists a polynomial-time estimator $\hat \mu(X_1, \dots, X_n)$ such that, with probability $\geq 1 - \beta$,  
    \[\|\hat \mu - \mu \|_2 \leq O\left(\sqrt{\frac{d}{n}} + \sqrt{\frac{\log 1/\beta}{n}}\right). \]
\end{theorem}

\subsection{Connections with Contamination}
At this point, we have seen a number of parallels between robust mean estimation under contamination and heavy-tailed settings.
In both cases, classic algorithms were either computationally inefficient or obtained a sub-optimal error. 
We could design efficient algorithms by defining an appropriate centrality notion (either spectral or combinatorial, respectively) and solving tractable optimization problems which either certify that a solution is good, or provide an ``interesting'' direction $v$ containing many outliers.
It is natural to ask whether there are deeper connections between the two problems. 

In fact, though not a focus of the original works \new{(and first observed by~\cite{DepersinL22})}, it is now \new{well understood} that with a careful setting of parameters, the combinatorial center in the median-of-means paradigm is \emph{automatically} contamination robust, thus giving both types of robustness simultaneously!
To see this, recall that our goal was to find a $(0.01, O(r^2_\beta))$-combinatorial center of the bucket means. 
This notion is naturally contamination robust: if (say) $0.001k$ of the bucket means were modified, then the ``centrality'' (i.e., the score function of~\eqref{prog:qp}) of any candidate center $\nu$  changes by at most $0.001k$.
Therefore, if we find a $(0.01, O(r^2_\beta))$-combinatorial center under such contamination, it was (at worst) a $(0.011, O(r^2_\beta))$-combinatorial center prior to contamination. 
Such loosening of the parameters turns out to be OK throughout the analysis, in particular for the relaxed SDP~\eqref{prog:sdp}.

However, we were reasoning about the number of \emph{bucket means} that were modified, and not the amount of contamination of the original dataset.
It is easy to relate these two quantities: if $\eta n$ (the number of contaminated points in the original dataset under $\eta$-strong contamination) is at most $0.001 k$, then no more than $0.001 k$ of the bucket means can be modified. 
This introduces the condition $k = \Omega(\eta n)$, in addition to the previous condition that $k = \Omega(\log 1/\beta)$.
Carrying through the analysis gives an estimator which is both contamination-robust and enjoys sub-Gaussian rates.
\begin{theorem}
\label{thm:contamination-subgaussian}
    Let $X_1, \dots, X_n$ be an $\eta$-strong contamination of $n$ samples from a distribution $\cD$ with mean $\mu \in \R^d$ and covariance $\Sigma \preceq I$. 
    There exists a polynomial-time estimator $\hat \mu(X_1, \dots, X_n)$ such that, with probability $\geq 1 - \beta$,  
    \[\|\hat \mu - \mu \|_2 \leq O\left(\sqrt{\frac{d}{n}} + \sqrt{\frac{\log 1/\beta}{n}} + \sqrt{\eta} \right). \]
\end{theorem}
To hint at where the rate comes from, we can substitute $k = \Omega(\eta n + \log 1/\beta)$ into Lemma~\ref{lem:comb-cent} to show that the mean $\mu$ is a combinatorial center of the bucket means with the described radius.
Note that this incorporates both the optimal rate in terms of both contamination (cf.\ Theorem~\ref{thm:second-moment-robust}) and sub-Gaussianity.

Others have subsequently investigated algorithmic connections between contamination and sub-Gaussian rates~\cite{DepersinL22,PrasadBR20}.
In a 2020 work, Hopkins, Li, and Zhang~\cite{HopkinsLZ20} made an important conceptual contribution, relating the two solution concepts we have seen thus far.
Specifically, they showed that the spectral center (Definition~\ref{def:spectral-center}) and combinatorial center (Definition~\ref{def:combinatorial-center}) are \emph{equivalent} to each other, up to some gap in constants (see Propositions 5.1 and 5.2 of~\cite{HopkinsLZ20}).
Algorithmically, this shows that one can use either the combinatorial \emph{or} the spectral center with the median-of-means paradigm to achieve sub-Gaussian rates.
\new{Simultaneously,} Diakonikolas, Kane, and Pensia~\cite{DiakonikolasKP20} showed \new{the more general result} that any contamination-robust estimator based on stability (of the sort used in Definition~\ref{def:stability}) can be employed as an aggregator in the median-of-means paradigm to achieve the sub-Gaussian rate.
\new{In fact, they further show that such estimators can be used \emph{directly} to achieve a near sub-Gaussian rate, entirely bypassing the median-of-means approach.
While this fact is conceptually interesting in its own right, it also allows one to employ higher-order moment information when establishing stability, leading to better contamination robustness than the $\sqrt{\eta}$ of Theorem~\ref{thm:contamination-subgaussian}.}

\section{Differential Privacy}
\label{sec:privacy}
As our last vignette, we turn a very different type of robustness: the celebrated notion of \emph{differential privacy}~\cite{DworkMNS06}.
\begin{definition}[\cite{DworkMNS06}]
    An algorithm $A\ :\ \mathcal{X}^n \rightarrow \mathcal{Y}$ is \emph{$(\varepsilon,\delta)$-differentially private (DP)} if, for all datasets $X, X' \in \mathcal{X}^n$ that differ in one entry and for all $S \subseteq \mathcal{Y}$, we have that
    \[\Pr[A(X) \in S] \leq e^\varepsilon \Pr[A(X') \in S] + \delta.\]
\end{definition}
This is a rigorous and quantitative notion of data privacy, which, informally speaking, prevents an adversary who observes the output of the algorithm from inferring too much information about its individual input datapoints.
A differentially private procedure is protected against a number of common risks, including regurgitation of input data, reidentification of individuals, dataset reconstruction, and more~\cite{DworkSSU17}.
Roughly speaking, $e^\varepsilon$ bounds the multiplicative increase in probability of any event if one person's data is included/excluded from the dataset, and $\delta$ bounds the additive increase in probability of the same, e.g., a catastrophic privacy loss.
For more discussion on the definition of differential privacy, see the book of Dwork and Roth~\cite{DworkR14}.

Differential privacy enforces that the estimator is robust, in a manner reminiscent of our discussion on contamination.
Both constraints require the estimator to behave nicely when elements of the dataset experience gross corruptions.
However, the superficial similarities end there, as the two settings differ in their parameter regimes (contamination-robust estimators typically tolerate corruptions of a \emph{constant fraction} of the dataset, whereas privacy pertains primarily to when \emph{one} point is corrupted),\footnote{Using a property known as \emph{group privacy}, differential privacy still gives meaningful guarantees when $\Theta(1/\varepsilon)$ points are changed. Nonetheless, observe that the two cases are distinguished in that estimators tolerate a constant \emph{fraction} versus a constant \emph{number} of modified points.} when their guarantees must hold (contamination-robustness only provides meaningful guarantees when the dataset is ``nice,'' whereas differential privacy must hold for \emph{every} dataset), and how strong these guarantees are (contamination-robustness only requires that the output is close, while privacy corresponds to the stronger condition that the \emph{distribution of} outputs is close).
Nonetheless, we will once again observe surprising technical and conceptual connections between contamination robustness and privacy.

\subsection{Privacy Fundamentals and Three Mean Estimators}

We will focus on private mean estimation under differential privacy, restricting our attention to distributions $\mathcal{D}$ with bounded covariance $\Sigma \preceq I$.\footnote{There is also significant study into private Gaussian mean estimation (see, e.g.,~\cite{KarwaV18,KamathLSU19,BunKSW19,BrownGSUZ21,BrownHS23,KuditipudiDH23,HopkinsKMN23}), but we focus on the bounded covariance setting which captures the most interesting difficulties.}
Given a dataset $X_1, \dots, X_n \in \R^d$, our goals for the estimator $\hat \mu(X_1, \dots, X_n)$ are twofold:
\begin{itemize}
    \item (Accuracy) If $X_1, \dots, X_n \sim \mathcal{D}$, $\|\hat \mu - \mu\|_2$ is small with high probability;
    \item (Privacy) For \emph{any} dataset $X_1, \dots, X_n$, $\hat \mu$ is $(\varepsilon, \delta)$-differentially private.\footnote{We will focus on the ``high privacy'' regime, where $\varepsilon \leq 1$.}
\end{itemize}

\new{We will first introduce three algorithms for private mean estimation, each deficient in a different way, and then describe an algorithm which overcomes these issues. 
The reader solely interested in connections with other forms of robustness may safely skip the first two.}

One of the simplest tools in differential privacy is the \emph{Laplace mechanism}.
Given a function, adding Laplace noise proportional to its $\ell_1$-sensitivity produces an output which is $(\varepsilon, 0)$-differentially private.
Recall that the (zero-mean) Laplace distribution with parameter $b$, denoted $\operatorname{Lap}(b)$, has PDF $\frac{1}{2b}\exp\left(-\frac{|x|}{b}\right)$ for all $x \in \R$.
\begin{proposition}[Laplace Mechanism]
    Let $f\ :\ \mathcal{X}^n \rightarrow \R^d$ be a function, and 
    \[\Delta_1^{(f)} \triangleq \max_{X, X' \in \mathcal{X}^n} \left\|f(X) - f(X') \right\|_1 \]
    be its $\ell_1$-sensitivity. 
    Then the Laplace mechanism is
    \[A(X) = f(X) + \left(Z_1, \dots, Z_d\right),\]
    where the $Z_i$ are independent $\operatorname{Lap}\left(\frac{\Delta_1^{(f)}}{\varepsilon}\right)$ random variables.
    The Laplace mechanism is $(\varepsilon,0)$-differentially private. 
\end{proposition}
Note that the Laplace mechanism gives a very strong guarantee of $(\varepsilon, 0)$-differential privacy.
This special case of $(\varepsilon,\delta)$-differential privacy with $\delta = 0$ is sometimes called \emph{pure} differential privacy.

We can use the Laplace mechanism to design an $(\varepsilon, 0)$-differentially private algorithm for mean estimation.

\begin{theorem}[\cite{KamathSU20}]
\label{thm:pure-dp-efficient-bad}
    Let $X_1, \dots, X_n$ be $n$ samples from a distribution $\cD$ with mean $\mu \in \R^d$ such that $\|\mu\|_2 \leq \sqrt{d}$ and covariance $\Sigma \preceq I$.\footnote{In the differential private setting, many estimators incur some dependence on a bound $R$ for the mean, i.e., a value such that $\|\mu\|_2 \leq R$. 
    In many cases this cost is logarithmic in $R$, and is shown to be necessary by matching lower bounds.
    To streamline our presentation, we focus on the case where we start with a ``coarse estimate'' for the mean $\mu$ with $\ell_2$-error $\sqrt{d}$ and re-center around it, yielding the condition $\|\mu\|_2 \leq \sqrt{d}$.}
    There exists a polynomial-time $(\varepsilon, 0)$-differentially private estimator $\hat \mu(X_1, \dots, X_n)$ such that, with probability $\geq 95\%$, 
    \[\|\hat \mu - \mu\|_2 \leq \tilde O\left(\sqrt{\frac{d}{n}} + \sqrt{\frac{d^{3/2}}{n\varepsilon}}\right).\footnote{In the parameter regime of interest (where $\varepsilon$ is small), the statistical error due to randomness of the samples $(\sqrt{d/n})$ is a lower-order term in the error rate.
Nonetheless, we leave it present in the expression to emphasize the cost of privacy.}\]
\end{theorem}
How can we use the Laplace mechanism to design a private algorithm for mean estimation?
The straightforward approach is to add Laplace noise to each coordinate of the empirical mean.
Unfortunately, the empirical mean has unbounded sensitivity, and thus the Laplace mechanism would prescribe adding infinite amounts of noise. 

The natural adjustment is to instead consider the \emph{clipped} mean: let $\operatorname{Clip}_\tau(X) = X$ if $\|X\|_2 \leq \tau$, and $\frac{\tau X}{\|X\|_2}$ otherwise.\footnote{Note that the clipped mean is itself a type of robust estimator. However, returning to the setting of contamination robustness, its cost of contamination would be the dimension-dependent $\eta\sqrt{d}$, falling short of the ideal $\eta$ that more sophisticated estimators obtained.
As we will see here, this weakly robust estimator will also be deficient for privacy purposes.}
That is, if a point has $\ell_2$-norm greater than $\tau$, we rescale it so that it sits on the $\ell_2$ ball of radius $\tau$. 
We can then apply the Laplace mechanism to this statistic: 
\[\frac{1}{n}\sum_{i=1}^n \operatorname{Clip}_\tau(X_i) + \operatorname{Lap}\left(\frac{\Delta_1^{(f)}}{\varepsilon}\right)^{\otimes d}.\]
The $\ell_1$-sensitivity $\Delta_1^{(f)}$ of the $\tau$-clipped mean can be computed to be $O\left(\frac{\tau}{n} \cdot \sqrt{d}\right)$: this is the \emph{$\ell_2$}-sensitivity of the statistic ($\tau/n$) scaled up by $\sqrt{d}$ to convert to a bound on the $\ell_1$-sensitivity.
The magnitude of the noise is thus $O\left(\frac{\tau \sqrt{d}}{n\varepsilon}\right)$ per coordinate, leading to the noise contributing an overall $\ell_2$-error of $O\left(\frac{\tau d}{n\varepsilon}\right)$.
On the other hand, the clipping operation itself introduces some error by \emph{biasing} the empirical mean. 
A calculation~\cite{KamathSU20} bounds the bias of clipping at $\tau$ by $O\left(\frac{\sqrt{d}}{\tau}\right)$.
Choosing the value of $\tau$ to balance the error due to bias and noise allows us to conclude Theorem~\ref{thm:pure-dp-efficient-bad}.

To highlight some key features of this algorithm, it:
\begin{itemize}
    \item is computationally efficient;
    \item enjoys the strong notion of pure $(\varepsilon, 0)$-differential privacy; and
    \item requires $n = \Omega(d^{3/2})$ samples to ensure constant error.
\end{itemize}
While the first two properties are favorable, the third property leaves something to be desired: recall that the non-private setting requires only $n = \Omega(d)$ samples to achieve a similar guarantee, and thus we have incurred an extra factor of $\Omega(\sqrt{d})$ in the requisite amount of data.
Our aim is to match the behavior in the non-private setting, and require only $n = \tilde O(d)$ samples to ensure constant error.

An slight variant of this approach appeals to another fundamental tool in differential privacy: the \emph{Gaussian} mechanism. 
The Gaussian mechanism adds noise scaled to the \emph{$\ell_2$-sensitivity} (which is smaller than the $\ell_1$-sensitivity) but guarantees only $(\varepsilon,\delta)$-DP with $\delta > 0$ -- this is sometimes referred to as \emph{approximate} differential privacy, and it is qualitatively weaker than pure DP.
A similar algorithm and analysis as before yields the following result.
\begin{theorem}[\cite{KamathSU20}]
\label{thm:approx-dp-efficient}
    Let $X_1, \dots, X_n$ be $n$ samples from a distribution $\cD$ with mean $\mu \in \R^d$ such that $\|\mu\|_2 \leq \sqrt{d}$ and covariance $\Sigma \preceq I$.
    There exists an polynomial-time $(\varepsilon, \delta)$-differentially private estimator $\hat \mu(X_1, \dots, X_n)$ such that, with probability $\geq 95\%$, 
    \[\|\hat \mu - \mu\|_2 \leq \tilde O\left(\sqrt{\frac{d}{n}} + \sqrt{\frac{d \sqrt{\log 1/\delta}}{n\varepsilon}}\right). \]
\end{theorem}

This algorithm:
\begin{itemize}
    \item is computationally efficient;
    \item satisfies approximate $(\varepsilon, \delta)$-differential privacy; and
    \item requires only $n = \tilde O(d)$ samples to ensure constant error.
\end{itemize}
This time, while the first and third properties are ideal, the second is deficient.
Though approximate DP is still a strong privacy notion, we would prefer the strongest notion of pure DP.

For yet another approach, we turn to a third tool in the differential privacy toolbox: the \emph{exponential mechanism}.
In contrast to the privacy mechanisms we have seen based on noise addition, the exponential mechanism is a \emph{sampling-based} algorithm.

\begin{proposition}[Exponential Mechanism~\cite{McSherryT07}]
\label{prop:exp-mech}
    Let $\mathcal{X}^n$ be the set of $n$ element datasets, $\mathcal{H}$ be a set of objects,  and $s\ :\ \mathcal{X}^n \times \mathcal{H} \rightarrow \R$ be a score function that outputs the ``quality'' of an object $h$ with respect to a dataset $X$. 
    Let 
    \[\Delta = \max_{h \in \mathcal{H}}\max_{X, X' \in \mathcal{X}^n} |s(X, h) - s(X',h)|\]
    be the sensitivity of the score function.
    
    The exponential mechanism is the algorithm which, given input dataset $X$, outputs an object $h \in \mathcal{H}$ with probability proportional to $\exp\left(\frac{\varepsilon s(X, h)}{2\Delta}\right)$.

    The exponential mechanism enjoys the following guarantees:
    \begin{itemize}
        \item (Privacy) It is $(\varepsilon, 0)$-DP with respect to the input dataset $X$
        \item (Utility) It outputs an object $h$ with score $s(X, h) \geq \operatorname{OPT}(X) - \frac{2\Delta}{\varepsilon}\left(\ln |\mathcal{H}| + t\right)$ with probability at least $1 - \exp(-t)$, where $\operatorname{OPT}(X) = \max_{h \in \mathcal{H}} s(X, h)$.\footnote{The exponential mechanism also applies to infinite $\mathcal{H}$, but we restrict our attention to the finite case for illustrative purposes.}
    \end{itemize}
\end{proposition}

The exponential mechanism assumes that some score function measures the quality (with respect to a dataset) of each object in a set.
The goal is to privately output an object with a large score. 
Non-privately, one would simply compute the score for all objects, and output the object with the largest score -- indeed, this corresponds to the exponential mechanism when $\varepsilon = \infty$.
The exponential mechanism instead samples an object randomly, with more probability assigned to objects with higher scores. 
The distribution is defined in a way that ensures pure differential privacy, and simultaneously gives very strong utility guarantees.

While the description of the exponential mechanism is somewhat abstract, it is quite a general and expressive primitive.
For example, the Laplace mechanism can be phrased as a special case: for a univariate function $f$, one can instantiate the exponential mechanism with $\mathcal{H} = \R$ and $s(X, h) = -|f(X) - h|$.

For many algorithmic tasks, the exponential mechanism serves as a simple and easy-to-analyze baseline, providing near-optimal error with a given amount of data.
The major caveat is that it is not, in general, computationally efficient. 
Naively, one must compute the score function for every object in $\mathcal{H}$ to define the probability distribution, and for many applications of interest, $\mathcal{H}$ will be exponentially large (or even infinite), making this computationally intractable.

With this caveat in mind, an algorithm based on the exponential mechanism provides the following guarantee.

\begin{theorem}[\cite{KamathSU20}]
\label{thm:pure-dp-inefficient}
    Let $X_1, \dots, X_n$ be $n$ samples from a distribution $\cD$ with mean $\mu \in \R^d$ such that $\|\mu\|_2 \leq \sqrt{d}$ and covariance $\Sigma \preceq I$.
    There exists an $(\varepsilon, 0)$-differentially private estimator\footnote{For a unified presentation, we describe a private estimator based on combinatorial centrality.
    This is thematically similar to (but technically different from) the algorithm in~\cite{KamathSU20}, which achieves the same rate.} $\hat \mu(X_1, \dots, X_n)$ such that, with probability $\geq 95\%$, 
    \[\|\hat \mu - \mu\|_2 \leq \tilde O\left(\sqrt{\frac{d}{n}} + \sqrt{\frac{d}{n\varepsilon}}\right). \]
\end{theorem}

Before we delve into details of the algorithm, we revisit the same three key properties of this algorithm:
\begin{itemize}
    \item it is not computationally efficient;
    \item it enjoys the strong notion of pure $(\varepsilon, 0)$-differential privacy; and
    \item it requires only $n = \tilde O(d)$ samples to ensure constant error.
\end{itemize}
While this algorithm achieves the ideal privacy and error guarantees, the computational intractability is problematic.

To instantiate the exponential mechanism, we need to define a set of objects and a score function.
Our objects will be candidates for the estimate of the mean: that is, a subset of $\Theta = \{\nu \in \R^d\ :\ \|\nu\|_2 \leq \sqrt{d}\}$.
To make our set of objects finite, we consider a \emph{cover} of $\Theta$, a subset $\mathcal{H} \subseteq \Theta$ such that for every $\nu \in \Theta$, there exists a nearby $\nu' \in \mathcal{H}$ such that $\|\nu - \nu'\|_2$ is small. 
Note that the size of such a cover is necessarily exponential in the dimension $d$.
Thus, naively, the running time of the exponential \new{mechanism} will also be exponential in $d$.
On the other hand, observe that the loss in utility (described by Proposition~\ref{prop:exp-mech}) has only a mild logarithmic dependence on the size of $\mathcal{H}$ (and thus a linear dependence on the dimension $d$) -- this will be important when bounding the final error. 

It remains only to define a score function.
The score function should indicate how ``good'' a candidate $\nu$ for the mean is, quality being measured with respect to the dataset $X$.
Furthermore, as prescribed by the utility guarantee in Proposition~\ref{prop:exp-mech}, the score function should also have low sensitivity.
These desiderata naturally lead us to revisit the quadratic program for combinatorial centrality \eqref{prog:qp}.\footnote{Revisited later, this score function can be seen as an instantiation of the inverse-sensitivity mechanism.}
Recall that a low objective function of \eqref{prog:qp} for a point $\nu$ indicates that it is a combinatorial center, and thus a good estimate for the mean.
Furthermore, since the objective function simply ``counts'' the number of far points, it is not hard to see that the sensitivity $\Delta$ is bounded by $1$. 
As a minor detail: since a combinatorial center corresponds to a point with a low objective function for \eqref{prog:qp} but the exponential mechanism tries to find a high scoring point, we actually consider the score function to be the \emph{negative} of the objective function of \eqref{prog:qp}. 

The more substantial difference in our application of \eqref{prog:qp} (in comparison to that in Section~\ref{sec:subgaussian}) is the number of pieces $k$ we partition our data into.
Recall that previously, we set $k = \Omega(\log 1/\beta)$ to achieve sub-Gaussian rates, and $k = \Omega(\eta n)$ to achieve robustness.
This time, we will set $k = \tilde \Omega(d/\varepsilon)$ for privacy.
Again, we split the dataset into $k$ parts and compute the bucket means of each, which we denote as $Y_1, \dots, Y_k$.
Following Lemma~\ref{lem:comb-cent}, we can see that the mean $\mu$ (and, to account for the discretization of the cover, any point which is sufficiently close) is a $\left(0.01, \tilde O\left(\frac{d}{\varepsilon n}\right)\right)$-combinatorial center \new{with high probability}.
This implies that the objective function of \eqref{prog:qp} for such a point will be $\geq 0.99k = \tilde \Omega(d/\varepsilon)$, and, by the utility guarantees of Proposition~\ref{prop:exp-mech}, the loss of score due to running the exponential mechanism will be only $O\left(\frac{\Delta}{\varepsilon}\log |\mathcal{H}|\right) = \tilde O \left(d/\varepsilon\right)$.
This implies that the exponential mechanism will still return a combinatorial center of the dataset, and consequently, a low-error estimate of the mean as described in Theorem~\ref{thm:pure-dp-inefficient}.
The algorithm is describe more precisely in Figure~\ref{alg:private-mean-slow}.

\begin{figure}
\caption{Algorithm for private mean estimation}\label{alg:private-mean-slow}
\begin{algorithmic}[1]
 \renewcommand{\algorithmicrequire}{\textbf{Input:}}
 \renewcommand{\algorithmicensure}{\textbf{Output:}}
\Require{Dataset $X = \{X_1, \dots, X_n\} \in \R^{n \times d}$, privacy parameter $\varepsilon$}
 \Ensure{Mean estimate $\hat \mu \in \R^d$}
 \Procedure{SlowPrivateMeanEstimation}{$X, \varepsilon$}
 \State $\mathcal{H} \gets$ an $\tilde O\left(\sqrt{\frac{d}{\varepsilon n }}\right)$-cover of the $\ell_2$-ball of radius $\sqrt{d}$
 \For{$i = 1$ to $k = \tilde \Theta(d/\varepsilon)$}
 \State $Y_i \gets \frac{k}{n}\sum_{j=1}^{n/k} X_{(i-1)\cdot\frac{n}{k} + j}$
 \EndFor
 \For{$\nu \in \mathcal{H}$}
 \State $s(X, \nu) \gets $ negative of the objective function of \eqref{prog:qp} on input $Y_1, \dots, Y_k$, $\lambda = \tilde O\left(
 \frac{d}{\varepsilon n}\right)$
 \EndFor
 \State sample $\hat \mu$ according to the distribution $p(\nu) \propto \exp\left(\frac{\varepsilon s(X,\nu)}{2}\right)$  for all $\nu \in \mathcal{H}$
 \State \Return{$\hat \mu$} 
 \EndProcedure
\end{algorithmic}
\end{figure}

\paragraph{A privacy trilemma} To summarize, at this point, we have seen three different algorithms, each of which suffers from one of these three deficiencies:
\begin{enumerate}
    \item computational intractability;
    \item approximate $(\varepsilon, \delta)$-differential privacy, rather than pure $(\varepsilon, 0)$-differential privacy; or
    \item requiring $n = \Omega(d^{3/2})$ samples to achieve constant error, rather than $n = \tilde O(d)$.
\end{enumerate}

\subsection{Efficient Multivariate Private Estimation}
In 2022, Hopkins, Kamath, and Majid~\cite{HopkinsKM22} gave the first efficient pure DP algorithm for mean estimation which requires only $n = \tilde O(d)$ samples to achieve constant error. 

\begin{theorem}[\cite{HopkinsKM22}]
\label{thm:pure-dp-efficient-good}
    Let $X_1, \dots, X_n$ be $n$ samples from a distribution $\cD$ with mean $\mu \in \R^d$ such that $\|\mu\|_2 \leq \sqrt{d}$ and covariance $\Sigma \preceq I$.
    There exists a polynomial-time $(\varepsilon, 0)$-differentially private estimator $\hat \mu(X_1, \dots, X_n)$ such that, with probability $\geq 95\%$, 
    \[\|\hat \mu - \mu\|_2 \leq \tilde O\left(\sqrt{\frac{d}{n}} + \sqrt{\frac{d}{n\varepsilon}}\right). \]
\end{theorem}
The algorithm is an adaptation of the aforementioned approach based on the exponential mechanism. 
This algorithm has the privacy and accuracy guarantees that we are aiming for, the only drawback is that it is not computationally efficient.
Recall that it is slow for two reasons: first, computing the score function for a single object is slow (since it involves solving the quadratic program \eqref{prog:qp}), and second, naively, we must compute the score function for exponentially many objects in our cover.
We will deal with these issues one by one.

To deal with inefficiency in computing the score function, we turn to the same ideas employed in Section~\ref{sec:subgaussian}.
Specifically, we can relax the QP \eqref{prog:qp} using the same SDP \eqref{prog:sdp} as before, and similarly adapt the analysis (e.g., bound the sensitivity, etc.). 

This might suffice if our goal were to efficiently and privately \emph{certify} a single candidate as a combinatorial center.
But since our goal is to efficiently and privately \emph{locate} such a point, we must further modify the framework.
Recall that optimization problems \eqref{prog:qp} and \eqref{prog:sdp} had two purposes: certifying that a point is a combinatorial center, and if not, selecting a direction that would progress towards finding one. 
However, the current invocation of the exponential mechanism only allows for the former. 
Similar to Section~\ref{sec:subgaussian} we adopt a descent-based approach, wherein each step, we use the exponential mechanism to select a \emph{direction} with high score (which here means that there are many points in that direction that are far). 
This doesn't solve our problems with sampling: if we took a cover over all directions, this would still be exponential in size, and computing the distribution for the exponential mechanism would remain intractable.

Instead, we appeal to the literature on efficient sampling from \emph{log-concave} distributions.
Indeed, the exponential mechanism samples from a distribution of the form $\exp\left(\varepsilon \cdot f(v)\right)$.
Thus, if $f(v)$ is concave, then the exponential mechanism samples from a log-concave distribution.
This turns out to be the case for the optimization problems we consider -- it is not hard to show this follows for \emph{any} linear function optimized over a convex set. 
Classical work on log-concave sampling generally guarantees that we sample from a distribution that is close is \emph{total variation distance} to the distribution of interest.
However, to provide $(\varepsilon, 0)$-differential privacy, we require multiplicative closeness on the probability of every event, which is stronger than total variation closeness.
Fortunately, Bassily, Smith, and Thakurta~\cite{BassilyST14} provide private log-concave samplers with precisely this style of bound.

Putting all the pieces together, the resulting algorithm ends up being very similar to the iterative descent-based algorithm that we used to obtain sub-Gaussian rates for heavy-tailed mean estimation.
The key difference is that, to guarantee privacy, every step is implemented ``noisily'' through the exponential mechanism.

More broadly, this algorithm is yet another example of using a combinatorial center in the median-of-means paradigm. 
Consequently, through judicious setting of parameters, this algorithm can simultaneously be near-optimally robust in all three senses we have considered so far!

\begin{theorem}
    Let $X_1, \dots, X_n$ be an $\eta$-strong contamination of $n$ samples from a distribution $\cD$ with mean $\mu \in \R^d$ such that $\|\mu\|_2 \leq \sqrt{d}$ and covariance $\Sigma \preceq I$. 
    There exists a polynomial-time $(\varepsilon, 0)$-differentially private estimator $\hat \mu(X_1, \dots, X_n)$ such that, with probability $\geq 1 - \beta$,  
    \[\|\hat \mu - \mu \|_2 \leq \tilde O\left(\sqrt{\frac{d}{n}} + \sqrt{\frac{d}{n\varepsilon}} + \sqrt{\frac{\log 1/\beta}{n \varepsilon }} + \sqrt{\eta} \right). \]
\end{theorem}
\subsection{More Connections with Contamination Robustness}

%We have seen how effective an algorithm from the mature field of robust statistics can be at addressing a problem in the (relatively) nascent field of private statistics. 
%It is natural to ask whether there are deeper connections at play.

We have observed that this algorithm, previously effective for robust mean estimation under contamination and heavy tails, is amenable to privatization.
While the former two types of robustness were closely linked, the technical connection with privacy established thus far has been relatively limited.
Indeed, the main common property employed for privacy and other forms of robustness is bounded sensitivity of the optimization problems~\eqref{prog:qp} and~\eqref{prog:sdp}. 
It is natural to ask whether deeper connections actually exist, or whether this application of a robust estimator was merely coincidental.
Though there is still much yet to be explored, there is good evidence that connections between contamination robustness and privacy are more fundamental in nature.

The first connections between contamination-robust and private statistics were explored in classic work by Dwork and Lei in 2009~\cite{DworkL09}.
They showed that many contamination-robust algorithms naturally lent themselves to privatization, via a celebrated framework they introduced known as propose-test-release (PTR).

Given the developments in algorithmic robust statistics described in this article, there has since been renewed interest in connections between contamination robustness and privacy. 
Many works borrow ideas and algorithms from the contamination-robustness literature, and show that careful adaptations can be made simultaneously contamination-robust \emph{and} private~\cite{BunKSW19,LiuKKO21,AshtianiL22,LiuKO22,HopkinsKM22,KothariMV22,AlabiKTVZ23} -- in many cases, providing the first (non-contamination-robust) private algorithm with the given guarantees for the problem, with contamination robustness as an added bonus.
Some of these works~\cite{KothariMV22,LiuKO22} provide general frameworks for adapting algorithms and analyses from the contamination-robustness literature to the private setting.

Perhaps the most broad and conceptual framework was introduced in simultaneous works by Hopkins, Kamath, Majid, and Narayanan~\cite{HopkinsKMN23} and Asi, Ullman, and Zakynthinou~\cite{AsiUZ23}. They showed that \emph{any} contamination-robust algorithm can be converted to a private one, in a completely black-box manner!
This is again done using the exponential mechanism, but a very specific form known as the \emph{inverse-sensitivity mechanism}.
The conversion is easy to describe.
We let the set of objects be the space of all possible parameters.
The score function of a candidate parameter is as follows: the (negative of the) minimum number of datapoints one must change so that a contamination-robust estimator's output is close to the candidate parameter.
Note that the sensitivity of this estimator is easily seen to be $1$, and that the highest scores (and thus the most weight) are assigned to candidate parameters close to the contamination-robust estimator's output.
The contamination-robustness property ensures that other candidate parameters with high scores don't stray too far away either. 

A significant caveat of this approach is that it is not computationally efficient.
Indeed, it is not clear how to compute the score function at all, even inefficiently.
However, Hopkins, Kamath, Majid, and Narayanan~\cite{HopkinsKMN23} further show that, for certain classes of contamination-robust estimators based on the Sum-of-Squares method, this conversion can be made algorithmic and computationally efficient.

The above discussion shows that contamination-robust estimators imply private ones.
Does the reverse hold?
It is \new{known} that private algorithms are \emph{automatically} contamination robust, with (small) contamination parameter $\eta \approx 1/\varepsilon \new{n}$.
This is \new{straightforward as a consequence of} a\new{n elementary} property of differential privacy known as \emph{group privacy}.
While differential privacy guarantees the similarity of an algorithm's output distributions on inputs at distance $1$, the group privacy property guarantees similarity on inputs at distance $k$, albeit with a quantitative degradation in similarity by a factor of $k$.
Setting $k \approx 1/\varepsilon n$, an easy calculation gives the desired result.
Asi, Ullman, and Zakynthinou~\cite{AsiUZ23} build upon this fact to show that contamination robustness and privacy are equivalent for low-dimensional problems. 
Georgiev and Hopkins~\cite{GeorgievH22} further show that any private algorithm with \emph{very high success probability} can enjoy even stronger contamination robustness, up to the parameter regime $\eta = \Omega(1)$.
Note that this type of high probability guarantee is reminiscent of the sub-Gaussian rate we were aiming for in Section~\ref{sec:subgaussian}, hinting at even more connections.

\section{Conclusion}
We discussed mean estimation under three different notions of robustness: under contamination, with heavy-tailed data, and subject to privacy constraints. 
In each instance, we ran into tensions involving computational inefficiency and other problem-specific desiderata.
We repeatedly found that the empirical mean, an optimal estimator in the non-robust setting, fell short of achieving our goals.
Perhaps surprisingly, despite the apparent dissimilarities in these three settings, we found that the same algorithmic ideas and techniques were effective in each case. 
Developed over the last decade, these have led to the first computationally efficient robust estimators across a variety of settings. 
Beyond these algorithmic connections, we have seen that there are deeper conceptual links between these settings as well.
Many more interesting and surprising connections exist beyond the scope of this article, and even more are surely yet to be discovered.

\section*{Acknowledgments}
GK would like to thank Gavin Brown, Yeshwanth Cherapanamjeri, Samuel B.\ Hopkins, Mahbod Majid, Argyris Mouzakis, Lydia Zakynthinou, \new{and the anonymous reviewers} for their helpful comments while writing this article. 
GK is supported by a Canada CIFAR AI Chair, an NSERC Discovery Grant, and an unrestricted gift from Google.


\bibliographystyle{alpha}
\bibliography{biblio}

\end{document}